\documentclass[conf]{new-aiaa}
\usepackage[utf8]{inputenc}

\usepackage{graphicx}
\usepackage{subfig}
\usepackage[version=4]{mhchem}
\usepackage{siunitx}
\usepackage{longtable,tabularx}
\usepackage{algorithmic}
\usepackage{graphicx}
\usepackage{textcomp}
\usepackage{hyperref}
\usepackage{marginnote}
\usepackage{xcolor}
\usepackage{times}
\usepackage{fancyhdr,graphicx}
\usepackage{gensymb}
\usepackage[ruled,vlined]{algorithm2e}
\usepackage{tikz}

\include{pythonlisting}
\def\BibTeX{{\rm B\kern-.05em{\sc i\kern-.025em b}\kern-.08em
    T\kern-.1667em\lower.7ex\hbox{E}\kern-.125emX}}
    
\newcommand{\Lone}{\mathcal{L}_1}

\newcommand{\IR}{\mathbb{R}}

\newcommand{\IZp}{\mathbb{Z}_{\geq0}}

\newtheorem{rem}{Remark}

\DeclareMathOperator{\atantwo}{atan2}

\usepackage[normalem]{ulem}

\newcommand\Mo[2]{{\color{orange}#1} {\color{blue} #2}}

\renewcommand\Mo[2]{{#2}}

\newcommand\replace[2]{{\color{red}\sout{#1}} {\color{orange}Suggestion:} {\color{blue} #2}}

\renewcommand\replace[2]{{#2}}

\newcommand\remove[2]{{\color{orange}Removed because:} {\color{orange} #1} {\color{red}\sout{#2}}}
\renewcommand\remove[2]{}

\newcommand\replaceVS[2]{{\color{red}\sout{#1}} {\color{orange}new name:} {\color{blue} #2}}

\renewcommand\replaceVS[2]{{#2}}

\title{Multi-level Adaptation for Automatic Landing with Engine Failure under Turbulent Weather}

\author{Haotian Gu\footnote{Ph.D. Student, Department of Mechanical Engineering, Hoboken, New Jersey, USA,  AIAA Student Member.} and Hamidreza Jafarnejadsani\footnote{Assistant Professor, Department of Mechanical Engineering, Hoboken, New Jersey, USA,  AIAA Member.}}
\affil{Stevens Institute of Technology, Hoboken, New Jersey, 07030}

\begin{document}

\maketitle

\begin{abstract}

This paper addresses efficient feasibility evaluation of possible emergency landing sites, online navigation, and path following for automatic landing under engine-out failure subject to turbulent weather. The proposed Multi-level Adaptive Safety Control framework enables unmanned aerial vehicles (UAVs) under large uncertainties to perform safety maneuvers traditionally reserved for human pilots with sufficient experience. In this framework, a simplified flight model is first used for time-efficient feasibility evaluation of a set of landing sites and trajectory generation. Then, an online path following controller is employed to track the selected landing trajectory. We used a high-fidelity simulation environment for a fixed-wing aircraft to test and validate the proposed approach under various weather uncertainties. For the case of emergency landing due to engine failure under severe weather conditions, the simulation results show that the proposed automatic landing framework is robust to uncertainties and adaptable at different landing stages while being computationally inexpensive for planning and tracking tasks. 

\end{abstract}

\section{Introduction}\label{section:Intro}



The unmanned aerial vehicles (UAVs) technology, which is moving towards full autonomous flight, requires operation under uncertainties due to dynamic environments, interaction with humans, system faults, and even malicious cyber attacks. Ensuring security and safety is the first step to making the solutions using such systems certifiable and scalable. In this paper, we introduce an autopilot framework called ``Multi-level Adaptive Safety Control" (MASC) for the resilient control of autonomous UAVs under large uncertainties and employ it for engine-out automatic landing under severe weather conditions.

\subsection{MASC Architecture}
 
In 2009, an Airbus A320 passenger plane (US Airways flight 1549) lost both engines minutes after take-off from LaGuardia airport in New York City due to severe bird strikes \cite{Sullenberger}. Captain Sullenberger\remove{does not serve anything w/o having Virtual Sully:}{(``Sully")} safely landed the plane in the nearby Hudson River. Inspired by this story, we aim to equip UAVs with the capability of human pilots to determine if the current mission is still possible after a severe system failure. If not, the mission is re-planned so that it can be accomplished using the remaining capabilities. \replace{The Virtual Sully Autopilot will perform safe maneuvers that are traditionally reserved for human pilots.}{ This is achieved by the \replaceVS{Virtual Sully Autopilot}{proposed autopilot framework, MASC,} which is capable of performing safe maneuvers that are traditionally reserved for human pilots.}

\begin{figure}[h!]
  \centering
  \subfloat{\includegraphics[width=0.4\textwidth]{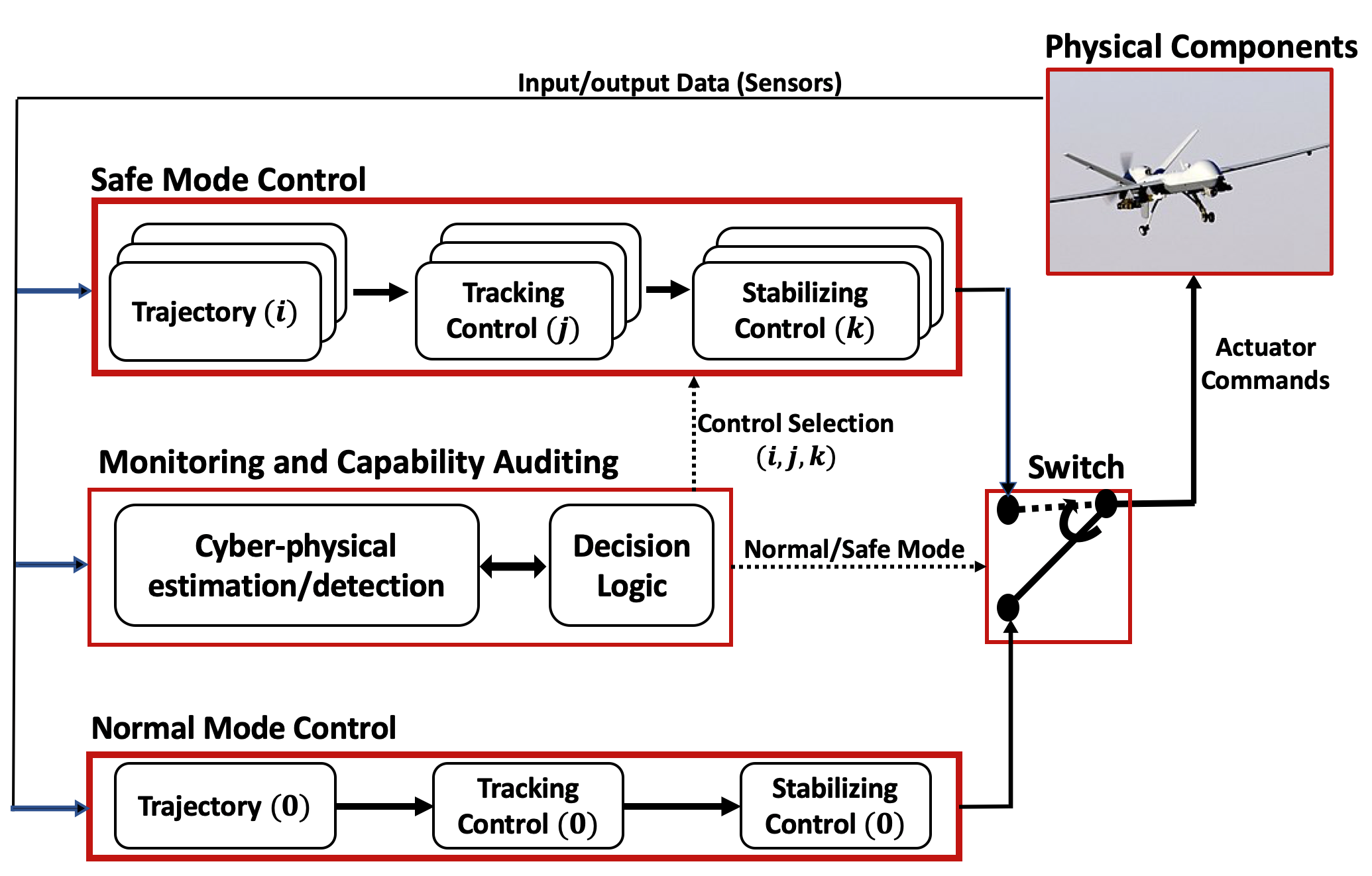}}
  \hfill
  \subfloat{\includegraphics[width=0.45\textwidth]{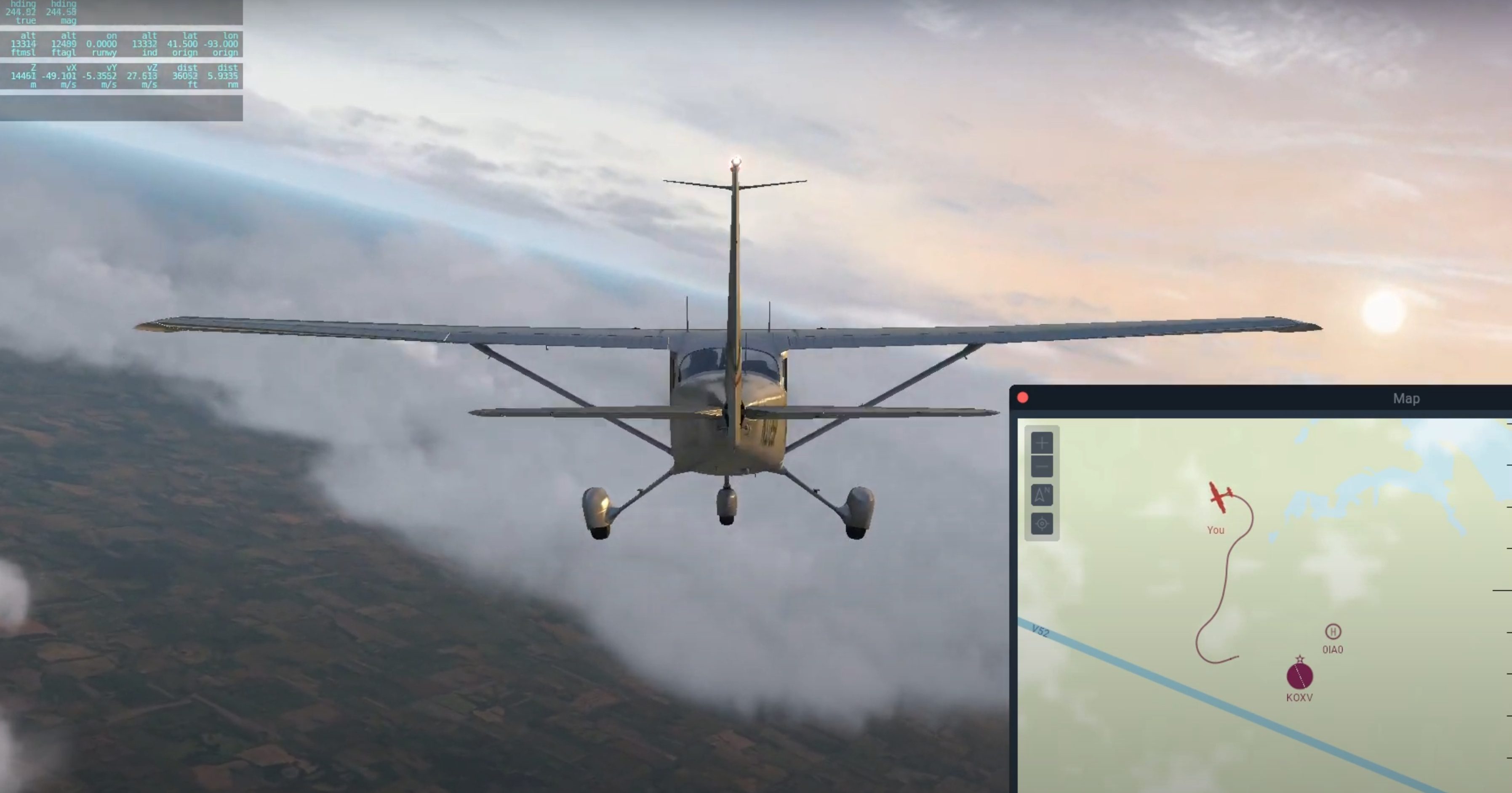}}
  \caption{\small (left) Multi-level Adaptive Safety Control (MASC) framework, (right) Simulation studies using X-Plane® program}\label{figure:autopilot_framework}
\end{figure}


\Mo{Structure:}{} From a mission control architecture perspective, we aim to replace the traditional top-down, one-way adaptation, \replace{between mission planning, trajectory generation, tracking, and stabilizing controller}{that starts with mission planning and cascades down to trajectory generation, tracking, and finally stabilizing controller}, with an \Mo{Add:}{integrated top-down and bottom-up architecture that allows for} two-way adaptation between planning and control to improve the stability and robustness of the system. To this end, we build the MASC framework upon the Simplex fault-tolerant architecture \cite{sha1998dependable,simplex2,crenshaw2007simplex,wang2013l1simplex,Yoon:VDRONE:2017}, which is recognized as a useful approach for the protection of cyber-physical systems against various software failures. By integrating the MASC framework with the Simplex architecture, we aim to enable cyber-physical systems to handle large uncertainties originating from the physical world. The \replaceVS{Virtual Sully Autopilot Framework}{MASC framework}, shown in Figure \ref{figure:autopilot_framework}, consists of the following components:

\begin{itemize}
    \item Normal Mode Controller: equipped with complex functionalities \replace{and operates}{to operate} the system under normal conditions. \Mo{Question: what complex functionalities?}{} 

    \item Safe Mode Controller with Multi-level Adaptation: a simple and verified controller that ensures safe and stable operations of the system with limited levels of performance and reduced functionalities. The control architecture consists of three levels: i) offline landing trajectory prediction; ii) \replace{trajectory regeneration and mission re-planning}{mission feasibility evaluation and re-planning}; and iii) online trajectory generation and path following control.
    
    \item Monitoring and Capability Auditing: uses a model considering the cyber-physical nature of autonomous systems for estimation and fault detection. The model identifies the remaining capabilities of the system and its decision logic triggers a \replace{Switch}{switch} from Normal Mode to Safe Mode.
\end{itemize}

Under the engine-out flight scenario with turbulent weather, the proposed architecture adapts the mission to the new constraints by i) auditing the remaining capability of the crippled aircraft and providing feedback to the other layers, ii) updating the flight envelope, iii) evaluating the feasibility of potential reachable destinations and selecting the low risk one; and iv) planning the flight path online and then employing a robust autopilot controller to track the path, meantime making sure to stay within the flight envelope of the crippled UAV. Feedback provided from the lower layers to the higher layers such as the mission planner allows for the interaction of the MASC modules and a two-way adaptation between the planning and control.


Emergency landing due to an engine failure under severe weather conditions is challenging even for an experienced pilot. Our proposed approach, referred to as the MASC framework, provides the autopilot with the agility required to compensate for uncertainties by adaptations in planning and control. The framework can be employed on dependable computing architectures for the safety control design. We utilize a computationally-efficient trajectory generation and tracking control approach, which can be computed with low latency on a low-cost real-time embedded platform onboard most UAVs. Also, the framework allows for the evaluation of reachable areas for an emergency landing. We tested the framework in a high-fidelity flight simulation environment for validation and verification. We achieved successful landings under a wide range of initial conditions (i.e., altitude, distance, and orientation relative to the landing site), windy weather, and turbulence without re-tuning the autopilot parameters.


\subsection{\Mo{Add:}{Related Work:} Engine-Out Emergency Landing}

UAVs play a significant role in a wide range of industries, including defense, transportation, and agriculture, to name a few \cite{Andrea_TASE_2014, puri2017agriculture, Lin_AJHP_2018}. Engine failure is one of the most hazardous situations for UAVs \cite{Jourdan_GNC_2010, ayhan2018path}. While engine failures are not common in passenger aircraft, an engine-out accident is more probable for a low-cost commercial UAV. The safety risks are even higher if a UAV crashes over a populated area endangering people and infrastructure on the ground. 
%
Numerous approaches for planning and control are proposed in the literature to mitigate the risks due to engine failure. In \cite{Atkins2006}, an adaptive flight planner (AFP) presented for landing an engine-out aircraft. The AFP approach for loss of thrust case performs the two main flight-planning tasks required to get a crippled aircraft safely on the ground: i) select a landing site and ii) construct a post-failure trajectory that can safely reach that landing site. An adaptive trajectory generation scheme with a certain presumed best glide ratio and bank angle for turns is proposed in \cite{atkins2010emergency}. Additionally, trajectory planning based on flight envelope and motion primitives is proposed in \cite{asadi2014damaged} for damaged aircraft. A reachable set for auto-landing is calculated by using optimal control theory in \cite{bayen2007aircraft}. 
Most of the related studies do not address emergency landing under additional weather uncertainties, and simulation results are mainly based on simplified models for the aircraft and environment.

The rapidly-exploring random trees (RRT) method is a popular sampling-based path planning algorithm designed to efficiently search nonconvex, high-dimensional spaces by randomly building a space-filling tree. The algorithm creates a search tree containing a set of nodes and the connecting path edges set. In \cite{Choudhury2013-hy}, a path planning scheme is developed based on the optimal sampling-based RRT algorithm to generate a landing trajectory in real-time and also examines its performance for simulated engine failures occurring in mountainous terrain. However, RRT-based algorithms are computationally demanding for planning large-scale smoother path~\cite{Slma2018EmergencyLG}. The demand increases with the dimensions of the searched state-space. Another motion planning for emergency landing is based on the Artificial Potential Field (APF) \cite{Chen2016-im} method and greedy search in the space of motion primitives. In APF~\cite{Dai2018-ap}, the UAV's path is calculated based on the resultant potential fields from the initial point to the target point. However, the conventional APF~\cite{Budiyanto2015-ot} may encounter the trap of a local minimum when the attractive force and repulsive force reach a balance, which means that the UAV stops moving towards the target.

\Mo{Airport Selection literature review:}


This paper is organized as follows. The components of the Multi-level Adaptive Safety Control (MASC) framework are presented in Section~\ref{section:MASC_Sec}. Particularly, monitoring and capability auditing is discussed in Section~\ref{subsection:Monitoring_and_Capability_Auditing}, the low-level safety controller is presented in Section \ref{subsection:Safe_Mode Control_with_Multi-Level_Adaptation}, and the mission adaptation is described in Section \ref{subsection:Mission}. Section~\ref{section:simulation_results} describes the high-fidelity software-in-the-loop (SITL) simulation environment for \replace{fixed-wing aircraft}{a fixed-wing aircraft} and presents the simulation results. Finally, Section \ref{section:conclusion} concludes the paper.



\section{Multi-level Adaptive Safety Control (MASC)}\label{section:MASC_Sec}

This section presents the components of the Multi-Level Adaptive Safety Control (MASC)\footnote{Open source code for the MASC framework on Github: (\url{https://github.com/SASLabStevens/MASC-Architecture.git})} framework.


\subsection{Monitoring and Capability Auditing}\label{subsection:Monitoring_and_Capability_Auditing}

The monitoring and capability auditing module has a set of stored expected models $\mathcal{D}=\{\mathcal{D}_1,\dots,\mathcal{D}_N\}$
where the triple
\begin{equation}
    \mathcal{D}_j=\{A_j,B_j,\Theta_j\}
    \label{eqn:dynamics}
\end{equation}
represents the plant matrices $(A_j,\,B_j)$, and the uncertainty set $\Theta_j$. 
In particular, each model is represented as
\begin{equation}
    \mathcal{D}_j: \left\{\begin{matrix}
    \dot{x}(t)  =  A_j x(t) + B_j(u(t) + f_j(x(t),t)),\\
    y(t)  =  C x(t), \quad x(t_0)=x_0,    \quad \quad \quad \quad \,\,
    \end{matrix}\right.
    \label{eq:System}
\end{equation} 
where $x(t) \in \mathbb{R}^n$ is the state vector, and $y(t) \in \mathbb{R}^q$ is the available output measurement.
The term $f_j \in \Theta_j,\,\, \forall (x,t) \in \mathbb{R}^n \times [0,\infty)$, represents unknown system uncertainties and disturbances subject to local Lipschitz continuity assumption.
Control input $u(t)$ is the robust low-level controller that stabilizes the model $\mathcal{D}_j$ with guaranteed robustness margins for \textit{apriori} given bounds on the uncertainties.
\begin{rem}
It is worth mentioning that $\mathcal{D}$ is a set of nominal/representative fault models. The models do not need to be perfectly accurate, and any modeling error is expressed as $f_j \in \Theta_j$. Given a nominal model, the safe mode controller will deal with any model mismatch or external disturbance.
\end{rem}

Monitoring and capability auditing is an integral part of the  \replaceVS{Virtual Sully Autopilot Framework}{MASC framework} (shown in Figure \ref{figure:autopilot_framework}), which performs the task of fault detection and isolation (FDI), i.e., to notice the existence of a fault, and to further identify the fault model. Since the autopilot is \Mo{add as a suggestion:}{characterized as} a cyber-physical system, regardless of the location of the faulty elements, the effect of the fault \Mo{add:}{is} always reflected in the physical world. Leveraging the measurement of physical state, we \replace{utilize}{employ} a model-based FDI approach used in control literature~\cite{hwang2009survey}\cite{Gao2015-ik}. Identifying the new model is critical for stabilizing the UAV, and it should be prioritized computationally, while the mission re-planning algorithm can take longer to converge to a feasible trajectory.

In the particular case of engine malfunction, monitoring the sensors such as the engine's RPM indicator and onboard accelerometer provides sufficient information for the monitoring and capability auditing module to detect the engine failure. Then, the module activates the planning and control task specifically designed for emergency landing within the Safe Control Mode module. Having identified the faults and updated the system model $\mathcal{D}_j$, another critical task of the monitoring and capability auditing module is to determine the safe flight envelope for the mission re-planing. Specifically, for protecting the flight envelope during the emergency landing, it is very crucial to maintain the forward airspeed around the optimal gliding speed $V_{\rm opt}$ and the corresponding best slope $\gamma_{\rm opt}$ that the aircraft manufacturer recommends. The optimal speed ensures maximum gliding distance without stalling the aircraft. In addition, the following constraints are considered for motion planning:
\begin{equation}
    \begin{split}
        V_{\rm min} &< V <V_{\rm max},\quad p_{\rm min} < p < p_{\rm max},\\
        \theta_{\rm min} &< \theta <\theta_{\rm max},\quad q_{\rm min} < q < q_{\rm max},\\
        \phi_{\rm min} &< \phi < \phi_{\rm max},\quad r_{\rm min} < r < r_{\rm max},\\
    \end{split}
    \label{eq:envelope}
\end{equation}
where $V$, $\theta$, and $\phi$ are the forward airspeed, pitch angle, and roll angle, respectively. Also, $p$, $q$, and $r$ are roll, pitch, yaw rates, respectively. 

\subsection{Safe Mode Path-Following Controller}\label{subsection:Safe_Mode Control_with_Multi-Level_Adaptation}

Path following controller modifies the control commands to the low-level longitudinal and lateral controllers to follow the reference path. Monitoring and Capability Auditing will compute the mission feasibility, given the states and the model of a damaged UAV. Large uncertainty mitigation requires mission adaptation and selection of a new trajectory that is still feasible, given the remaining capabilities.
For large uncertainties outside the design bounds, i.e., $f_j \not\in \Theta_j$ \Mo{add:}{in \eqref{eq:System}}, the control inputs can saturate and drive the system to unsafe states. Modification of the reference command $r_{\rm d}[i]$\remove{}{ in \eqref{eq:control}} based on the updated objectives is another layer of defense for maintaining safety by {\em satisfying flight envelope constraints}. Therefore, we consider a control structure that consists of a path following controller, where the generated reference commands to the low-level controller are limited by saturation bounds to maintain the closed-loop system within operational safety envelope.

Let the reference command be constrained to a convex polytope as a safe operational region, defined by the set
\begin{equation}\label{eq:safety_envelope}
\mathcal{R} = \left\{ {r_{\rm d} \in {\IR^q}|\,{{\left\| {Wr_{\rm d}} \right\|}_{\infty }} \le 1} \right\},
\end{equation}
where $W =\rm{diag}\{r_{{\rm{max}}_1}^{-1},\,...,r_{{\rm{max}}_q}^{-1}\}$, and the positive constants $\rm{r}_{{\rm{max}}_i}$'s are the saturation bounds on the reference commands. Then, the weighted reference command is bounded by
    \[\left\|Wr_{\rm d}[i]\right\|_{\infty} \le 1,\quad i\in\IZp.\] 
In this paper, the reference command $r_{\rm d}[i],\,\, i \in \IZp$, which is generated by the path following control law, is given by
{\small \begin{equation}\label{eq:reference_command}
\begin{split}
r_{\rm d}[i] &=(1-\alpha)W^{-1}{\rm sat}\left\{\frac{1}{1-\alpha}WF_{\rm z}\left(z_{\rm{m}_{\rm d}}[i]-z_{\rm d}[i]\right)\right\},\\
\end{split}
\end{equation}
}where ${\rm sat}\{\cdot\}$ denotes the saturation function,  $F_{\rm z} \in \IR^{q \times p}$ is the state-feedback gain, and $\alpha \in (0,\,1)$ is a constant. Also, $z_{\rm m_ d}[i]$ is the desired trajectory variable generated by the mission planner and $z_{\rm d}[i]$ is the actual state of the aircraft. 
In the case of an emergency glide landing, the \Mo{as suggestion, add:}{desired} heading angle $\psi$ and flight path angle $\gamma$ are the variables that are generated by the mission planner. The path following controller in \eqref{eq:reference_command} is equivalent to a PI controller subject to a saturation function. This control law ensures that the roll and pitch commands stay within the safe flight envelope during the glide landing, and hence the possibility of a stall decreases.

\subsection{Mission Adaptation}\label{subsection:Mission}

In the  \replaceVS{Virtual Sully Autopilot Framework}{MASC framework}, we present a mission planner in the Safe Control Mode that generates the landing trajectory to a safe landing site. The planner also evaluates mission feasibility considering the damage of the aircraft and environmental constraints. 
 \Mo{Add:}{In the case of a fault/failure detection,} once the capability auditing provides the new/altered model $\mathcal{D}_j$\remove{}{ to the MASC}, two concurrent steps will be taken: {\em i)} MASC initiates the feasibility evaluation and landing trajectory estimation, and {\em ii)} autopilot controller is activated based on its ability to stabilize the system around a pre-calculated
\replace{$h^{th}$}{$r^{th}_{\rm d}[\cdot]$} reference command. The first step is taken to ensure that the system does not violate its stability/safety bounds while the mission is being re-planned. Once the mission is re-planned, it is fed into the path following controller, which then accordingly alters the \replace{$h^{th}$}{$r^{th}_{\rm d}[\cdot]$} reference command provided to the $j^{th}$ low-level controller to execute the new mission. 
\replace{
In this paper, we assume that the coordinates and the direction of the safe landing site have already been computed and focus on the landing under the severe weather condition.

In future work, we will enable the mission re-planning to compute the {\it reachable area}, which is defined as all the spatial points the aircraft is capable of reaching given the dynamic constraints, potential and kinetic energy, and available fuel if the engine is still partially working. Also, the mission planner will be able to change the destination as conditions change.}{In the second step, re-planning of the mission, it is crucial to compute the {\it reachable area}, which is defined as all the spatial points the aircraft is capable of reaching given the dynamic constraints, potential and kinetic energy, and available fuel if the engine is still partially working. To this end, provided the information and new constraints by the monitoring and capability auditing module, a set of candidate locations are initially considered. This initial set may include nearby airports and empty lands. Leveraging the updated model of the UAV, MASC evaluates the feasibility of safe landing in the candidate areas and identifies the most likely safe location for an emergency landing. Any violation of the safety constraints (for instance \eqref{eq:envelope} and \eqref{eq:safety_envelope}) during the evaluation process rules out a candidate area as a safe reachable area.} 

%

\begin{figure}[h!]
    \centering
    \includegraphics[width=0.7\textwidth]{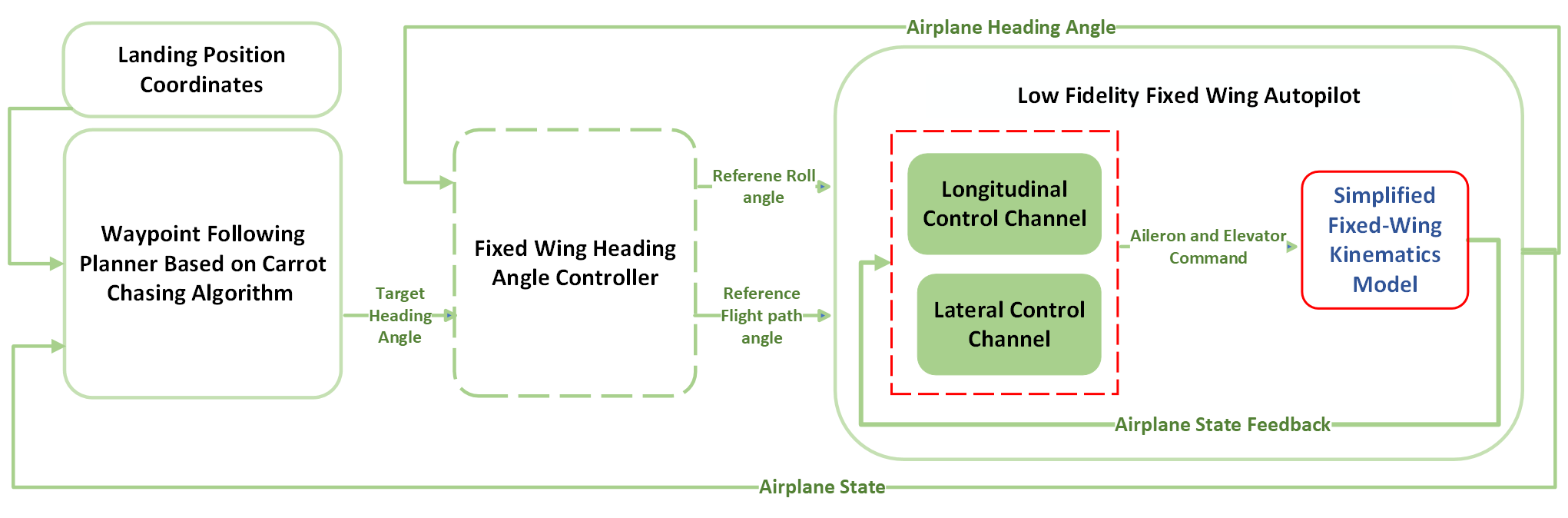}
	\caption{\small The block diagram for  evaluating feasibility of the landing trajectories using a simplified UAV model in Matlab}
	\label{Fig:Framework of waypoint follower by software in the loop simulation}
    \end{figure}

The feasibility evaluation for landing site selection is summarized in Algorithm~\ref{algo:evaluation}, and the landing trajectory planning is summarized in Algorithm~\ref{algo:carrot}. The feasibility evaluation can be viewed as an offline trajectory planning from the initial state of the engine-out aircraft to a few possible landing coordinates. For the offline trajectory generation, a simplified and reduced model of the UAV dynamics and autopilot is used to estimate the trajectory and travel time to each landing site. Consequently, the more desirable landing coordinate is selected, and it is passed to the online trajectory planning for the online execution. The block diagram for feasibility evaluation of the landing trajectories using a simplified UAV model is shown in Figure~\ref{Fig:Framework of waypoint follower by software in the loop simulation}. Both online and offline planning use the carrot chasing based guidance logic for trajectory following \cite{sujit2014unmanned}. The carrot chasing method uses a pseudo target moving along the desired flight path while generating the desired heading angle using the reference point~\cite{Park2004}, and it is robust to disturbances~\cite{Sujit2013-im}.

\begin{figure}
    \begin{center}
        \begin{tikzpicture}[scale=0.15]
        \tikzstyle{every node}+=[inner sep=0pt]
        \draw [black] (23.7,-36.8) circle (3);
        \draw (23.7,-36.8) node {I};
        \draw [black] (40.1,-36.8) circle (3);
        \draw (40.1,-36.8) node {II};
        \draw [black] (56.1,-36.8) circle (3);
        \draw (56.1,-36.8) node {III};
        \draw [black] (41.423,-39.48) arc (54:-234:2.25);
        \fill [black] (38.78,-39.48) -- (37.9,-39.83) -- (38.71,-40.42);
        \draw [black] (25.023,-39.48) arc (54:-234:2.25);
        \fill [black] (22.38,-39.48) -- (21.5,-39.83) -- (22.31,-40.42);
        \draw [black] (57.423,-39.48) arc (54:-234:2.25);
        \fill [black] (54.78,-39.48) -- (53.9,-39.83) -- (54.71,-40.42);
        \draw [black] (26.7,-36.8) -- (37.1,-36.8);
        \fill [black] (37.1,-36.8) -- (36.3,-36.3) -- (36.3,-37.3);
        \draw [black] (26.131,-35.045) arc (122.47267:57.52733:25.645);
        \fill [black] (53.67,-35.05) -- (53.26,-34.19) -- (52.73,-35.04);
        \draw [black] (43.1,-36.8) -- (53.1,-36.8);
        \fill [black] (53.1,-36.8) -- (52.3,-36.3) -- (52.3,-37.3);
        \end{tikzpicture}
    \end{center}
    \caption{\small Landing mission phase transition diagram. \Mo{add:}{Phase I: Cruising, Phase II: Loitering, and Phase III: Approach.}}
    \label{figure:phase_state_machine}
\end{figure}
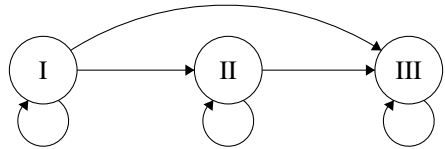


The landing mission is divided into three phases: \textit{Phase I: Cruising, Phase II: Loitering, and Phase III: Approach} as illustrated in Figure \ref{figure:phase_state_machine}. In Algorithm \ref{algo:carrot}, the emergency glide landing procedure starts by cruising towards the loitering center with coordinates $(x_{\rm l},\,y_{\rm l})$ near the landing site. After the aircraft is close enough to the loitering center, i.e., $\sqrt{(x_{\rm l}-x)^2+(y_{\rm l}-y)^2} <R_{\rm l}$, the mission enters the loitering phase. While loitering, the aircraft losses altitude in a spiral trajectory. When the cut-off altitude, $z_{a}$, is reached, i.e., $z<z_{a}$, the mission enters the approach phase. Notice that the mission is allowed to progress in one direction, similar to a directed acyclic graph (DAG), and it is possible that the first and second phases are skipped altogether depending on the initial states of the aircraft, as illustrated in Figure \ref{figure:phase_state_machine}. The variables defined for the three flight phases in Algorithm\ref{algo:carrot} are described in the following.

{\it Phase I:} During the cruising phase, the aircraft cruises towards the loitering circle. The desired heading angle is calculated as given in Algorithm \ref{algo:carrot}. The line-of-sight (LOS) $\theta$ denotes the angle in the Euclidean plane, given in radians, between the positive $x$-axis and the ray to the point $(W_{\rm ipl},\,W_{\rm i})$ which connects the initial engine malfunction position to the loiter center. $ path_{\rm g}$ is the Euclidean distance of the current coordinates and the start of the reference path. $R_{\rm u}$ is the distance between the start of the reference path and the current particle projected on the straight reference path. The airspeed should remain close to the best gliding speed $V_{\rm opt}$ while the UAV cruises to the landing site. 

{\it Phase II:} In the loitering phase, the aircraft loiters near the landing site in a spiral trajectory to lose any excessive altitude for the final approach. In Algorithm~\ref{algo:carrot}, $O$ denotes the global coordinates of the loiter center. Also, $\theta$ will gradually approach the chord tangent angle of loiter as the moving trajectory converges to the reference circle. In addition, $\delta$ is the look-ahead distance. With an increase in $\delta$, the flight path of the UAV can converge to the reference trajectory quickly, reducing the cross-track error \cite{bemporad1999robust}.

{\it Phase III:} In the approach phase, the aircraft aligns itself with the runway and tracks the desired flight path angle for the final approach. The trajectory point renew scheme is the same as the Phase I. The difference is $\theta$ denotes the angle in the Euclidean plane, given in radians, between the positive $x$-axis and the ray to the points $(W_{\rm ipl},\,W_{\rm i})$ which connects optimal landing area coordinates to $(x_{\rm u},\,y_{\rm u})$. If the ground distance of the aircraft to the landing site is larger than a constant $R$, i.e., $\sqrt{(x-x_{\rm f})^2+(y-y_{\rm f})^2}>R$, we have
{\small
\begin{equation}\label{eq:starting point of reference point for phase III}
\begin{split}
x_{\rm u} = x_{\rm l} + R_{\rm l} \times cos(\psi_{\rm f} - \pi),\\
y_{\rm u} = y_{\rm l} + R_{\rm l} \times sin(\psi_{\rm f} - \pi),\\
\end{split}
\end{equation}
}
where $R_{\rm l}$ is loiter diameter, and $\psi_{\rm f}$ is the heading angle of the runway. Also, $(W_{\rm ipl},\,W_{\rm i})$ connects the $(x_{\rm u},\,y_{\rm u})$ start of the reference line and landing position coordinates $(x_{\rm f},\,y_{\rm f})$ in Euclidean plane. 

\begin{algorithm}
    \SetAlgoLined
    \textbf{Input:} Initial engine out coordinates, coordinates of the back up landing areas\
    
    \textbf{Output:} Predicted landing trajectory and estimated landing time\
    
    \textbf{Procedure For Feasibility Evaluation of Landing Areas:}\

     If(Engine malfunction == true)
     
    \setlength{\parindent}{40pt} MASC Framework initiates the Feasibility Evaluation process.
     
    \setlength{\parindent}{40pt} GO TO Landing area feasibility evaluation and landing trajectory and time prediction
    
    \setlength{\parindent}{40pt}1: Feed the global coordinates where engine is out and of landing areas;
    
    \setlength{\parindent}{40pt}2: Implement offline path planning in acceleration mode;
    
    \setlength{\parindent}{40pt}3: Get the predicted flight trajectories and estimated landing time;
    
    \setlength{\parindent}{40pt}4: Determine the most suitable landing site;
     
    \setlength{\parindent}{20pt}do 
    
    \setlength{\parindent}{40pt}MASC(Online Navigation) Initiate
    
    \setlength{\parindent}{20pt} while(The determined optimal landing coordinates)

    \setlength{\parindent}{0pt}else

    \setlength{\parindent}{40pt}  Conducting the Normal Flight mode;

     \caption{Algorithm for feasibility evaluation for candidate landing areas.}
     \label{algo:evaluation}
\end{algorithm}
\vspace{0mm}

\begin{algorithm}
    \SetAlgoLined
    \textbf{Input:} Initial Engine out coordinates, reachable landing coordinates and runway direction.\
    
    \textbf{Output:} Desired Heading Angle\
    
    \textbf{SIL simulation Procedure:}\
    
    if (Engine malfunction == true)\
    
\setlength{\parindent}{20pt} if ($Distance >R_{l} $ and   $z>z_{a}$)\

     \setlength{\parindent}{40pt}begin if
     
     \setlength{\parindent}{40pt} 1: Initialize: $W_{\rm i}=(x_{\rm i},\,y_{\rm i}),\,W_{\rm ipl}=(x_{\rm ipl},\,y_{\rm ipl}),\,P_{\rm new}=(x_{\rm new},\,y_{\rm new}),\,\psi_{\rm des},\, \delta,\,\theta\,$;
    
        2: $path_{\rm g}=\left\|W_{\rm i}-P_{\rm new}\right\|$;
    
        3: $\theta = \left\|W_{\rm ipl}-W_{\rm i}\right\|$;
        
        4: $\theta_{\rm u} = \left\|P_{\rm new}-W_{\rm i}\right\|$;
    
        5: $l_{\rm d}=\theta-\theta_{\rm u}$;\
    
        6: $R_{\rm u}= \sqrt{(path_{\rm g})^2+(path_{\rm g}\times\sin(l_{\rm d})))^2}$;\
    
        7: $(x_{\rm t},\,y_{\rm t})= ((R_{\rm u}+\delta)\times\cos(\theta),\,(R_{\rm u}+\delta)\times\sin(\theta))$;\
    
        8: $\psi_{\rm des}=\atantwo(y_{\rm t}-y_{\rm new},\,x_{\rm t}-x_{\rm new})$;
     
        \setlength{\parindent}{40pt}end if
        
     \setlength{\parindent}{20pt} else if ($Distance < R_{l}$ and $z>z_{a}$)\
     
        \setlength{\parindent}{40pt}begin if
        
        \setlength{\parindent}{40pt} 1: Initialize: $O=(x_{\rm l},\, y_{\rm l}),\,P=(x,\,y),\, \psi_{\rm des},\, \lambda,\, \kappa$;\
    
        2: $d=\left\|O-P\right\|-R_{\rm c}$;
    
        3: $\theta_{\rm l} = \text{atan2}\left( y-y_{\rm l},\, x-x_{\rm l}\right)$;
    
        4: $(x_{\rm t},\,y_{\rm t}) = (x_{\rm l}+R_{\rm c}\times\cos(\lambda+\theta_{\rm l}),\,y_{\rm l}+R_{\rm c}\times\sin(\lambda+\theta_{\rm l}))$;\
    
        5: $\psi_{\rm des}=\atantwo(y_{\rm t}-y,\,x_{\rm t}-x)$;

        \setlength{\parindent}{40pt}end if
        
   \setlength{\parindent}{20pt}else\
   
        \setlength{\parindent}{40pt}begin if
        
       \setlength{\parindent}{40pt} 1: Initialize: $W_{\rm i}=(x_{\rm i},\,y_{\rm i}),\,W_{\rm ipl}=(x_{\rm ipl},\,y_{\rm ipl}),\,P_{\rm new}=(x_{\rm new},\,y_{\rm new}),\,\psi_{\rm des},\, \delta,\,\theta$;
    
        2: $path_{\rm g}=\left\|W_{\rm i}-P_{\rm new}\right\|$;
    
        3: $\theta = \left\|W_{\rm ipl}-W_{\rm i}\right\|$;
        
        4: $\theta_{\rm u} = \left\|P_{\rm new}-W_{\rm i}\right\|$;
    
        5: $l_{\rm d}=\theta-\theta_{\rm u}$;\
    
        6: $R_{\rm u}= \sqrt{(path_{\rm g})^2+(path_{\rm g}\times\sin(l_{\rm d})))^2}$;\
    
        7: $(x_{\rm t},\,y_{\rm t})= (R_{\rm u}+\delta)\times\cos(\theta),\,(R_{\rm u}+\delta)\times\sin(\theta))$;\
    
        8: $\psi_{\rm des}=\atantwo(y_{\rm t}-y_{\rm new},\,x_{\rm t}-x_{\rm new})$;
        
        \setlength{\parindent}{40pt}end if
        
    \noindent else 
    
       \setlength{\parindent}{40pt} Fly in Normal Mode;\
        
    \noindent end;

     \caption{Algorithm for emergency landing using the nonlinear guidance logic \cite{sujit2014unmanned}.}
     \label{algo:carrot}
\end{algorithm}
\vspace{0mm}

%

%

\section{Software-in-the-loop Simulation Study}\label{section:simulation_results}



\begin{figure}[h!]
    \centering
    \includegraphics[width=0.7\textwidth]{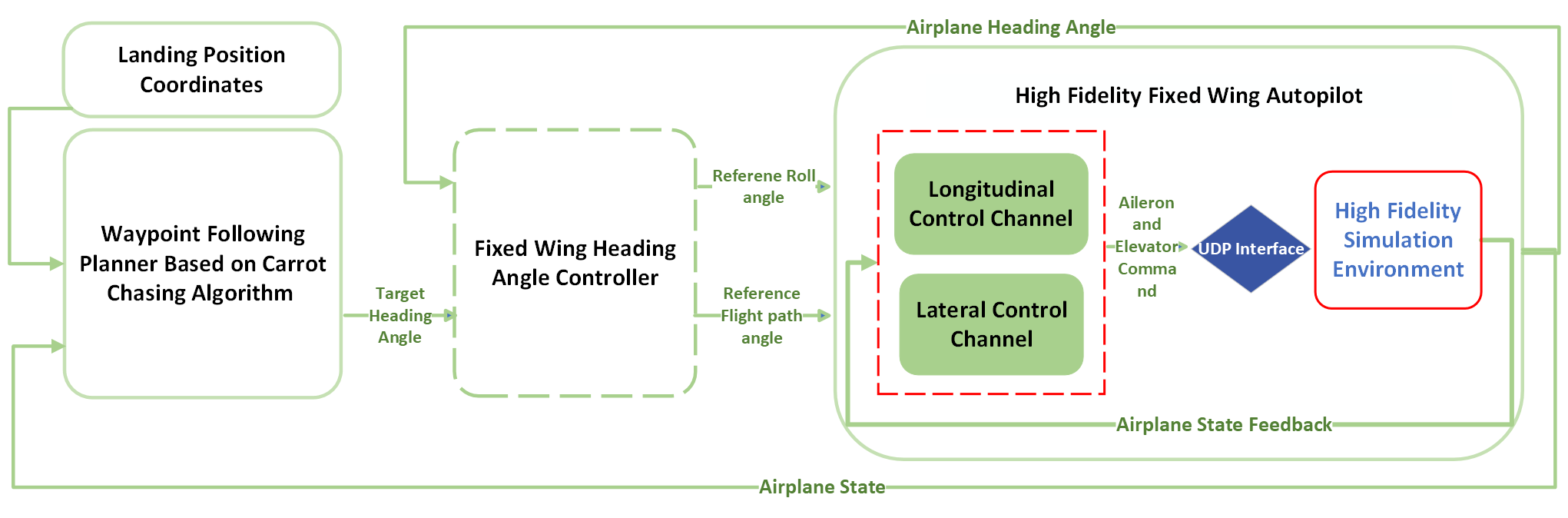}
	\caption{\small  The block diagram for implementing MASC framework in the X-Plane high-fidelity flight simulation environment}
	\label{Fig:MASC autopilot}
    \end{figure}

This section presents a software-in-the-loop (SITL) simulation scheme to evaluate and validate the proposed MASC\footnote{Open source code for the MASC framework on Github: (\url{https://github.com/SASLabStevens/MASC-Architecture.git})} framework for online path planning and navigation under the emergency landing case. The SITL architecture for the MASC autopilot, shown in Figure \ref{Fig:MASC autopilot}, has three primary components: i) high-fidelity physical simulation environment (in X-Plane), and ii) MASC autopilot (in MATLAB/Simulink) consisting of a nonlinear logic based mission planner and proportional heading angle regulation scheme, and iii) a user datagram protocol (UDP). The X-Plane program has been certified by the Federal Aviation Administration (FAA) as a simulation software to train pilots. The aircraft model used in this simulation is Cessna 172SP as shown in Figure~\ref{figure:autopilot_framework}.


\subsection{UDP Receiver and Sender Interface}


X-Plane adopts the User Datagram Protocol (UDP) to communicate with the third-party software and external processes. Unlike the Transmission Control Protocol (TCP), UDP assures that data packages will arrive completely and orderly. Also, the communication via UDP can achieve high-speed data traffic compared to other protocols by efficient use of bandwidth, which is an advantageous characteristic for the simulation under consideration. Correspondingly, DSP toolbox in MATLAB/Simulink supports the UDP communication through DSP System Toolbox. This toolbox can query an application using UDP to send real-time data from the Simulink model to the corresponding channel in X-Plane. Also, the UDP object allows performing byte-type and datagram-type communication using a UDP socket in the local host. For the implementation, we consider designing subscriber and publisher to guarantee data transfer between the X-Plane and MATLAB/Simulink in real time. For subscriber, we used two Simulink blocks: an embedded MATLAB function and a byte unpack. For publisher, we use a byte pack and encoder, and both are linked via a bus module in Simulink.

\subsection{Engine-Out Landing under Clear Weather}

\begin{figure}[h!]
    \centering
    \includegraphics[width=1\textwidth]{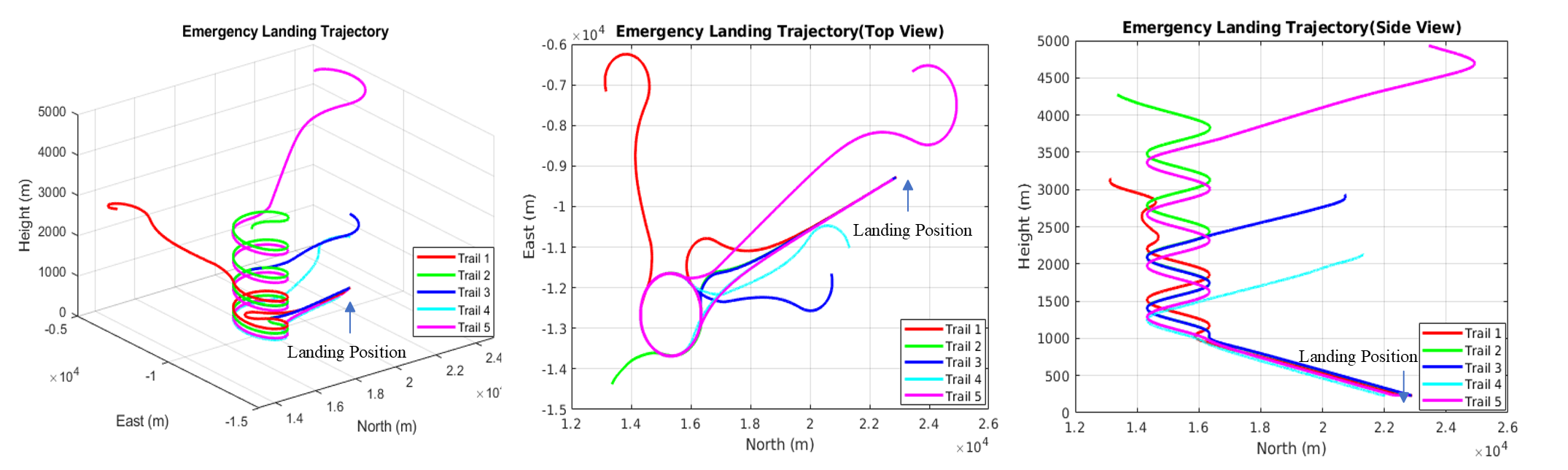}
	\caption{\small Automatic engine-out landing under clear weather from different initial positions (simulation video \href{https://www.youtube.com/watch?v=d_yzGMl9LYs}{link})}
	\label{Fig:Simulation I}
    \end{figure}

The simulation results for the automatic landing process under clear weather are shown in Figure~\ref{Fig:Simulation I}, illustrating three different viewpoints of the real-time landing trajectory. In this emergency landing simulation, we randomly initiate the process from five different positions given in Table~\ref{tab:Different starting positions with engi}. The final landing coordinates are set to North $21822m$, East $-9751.8m$, Height $140m$, and  Heading direction of the runway $24.18 deg$. As we can see from the results, the MASC for emergency landing during engine failure can plan a path online to safely navigate the aircraft to the configured landing site from various initial conditions. Also, the aircraft's airspeed remains well above the stall speed, which is around 30 $m/s$. Throughout the simulations, the weather condition is set to Clear, with no wind, i.e., the best weather condition in X-Plane®.

\begin{table}[ht]
\centering
\begin{tabular}{ | m{2.3em} | m{2.3cm}| m{2.3cm} |m{2.3cm} |m{2.3cm} | } 
  \hline
  Trial& North ($m$) & East ($m$) & Height ($m$) & Heading ($Deg$)\\
  \hline
  1 & 13163 & -7164.9 & 3000 & 78.5 \\ 
  \hline
  2 & 13353 & -14380 & 4000 & 110.3 \\ 
  \hline
  3 & 23429 & -6675.6 & 5000 & 85.6 \\
  \hline
  4 & 20719 & -11652 & 3000 & 256.74 \\
  \hline
  5 & 21323 & -11021 & 2000 & 69.594\\
  \hline
\end{tabular}
\caption{\small The initial position coordinates of the engine-out aircraft in the simulation trials}
\label{tab:Different starting positions with engi}
\end{table}

\subsection{Engine-Out Landing under Windy and Turbulent Weather} 

\begin{figure}[h!]
    \centering
    \includegraphics[width=1\textwidth]{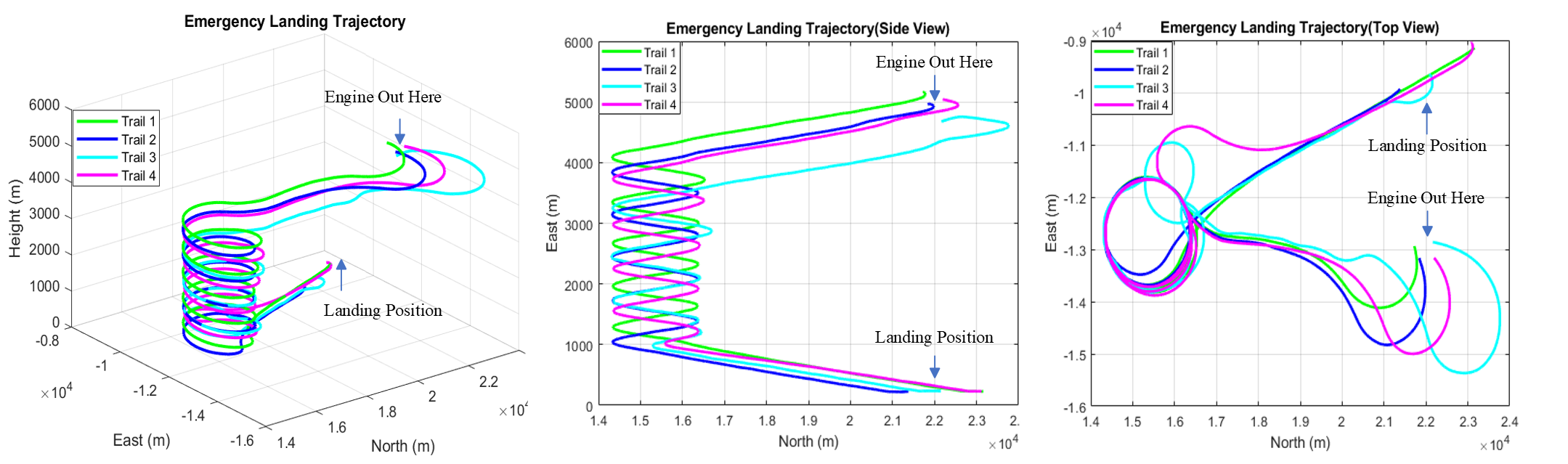}
	\caption{\small Automatic engine-out landing under different windy and turbulent weather settings in the X-plane program (simulation video \href{https://www.youtube.com/watch?v=FpEaSQYd7Jw}{link})}
	\label{Fig:Simulation II}
    \end{figure}

    
To further demonstrate the robustness of the MASC framework under large wind and turbulence uncertainties, we simulated the emergency landing task under different severe weather settings in the X-Plane program with parameters listed in Table \ref{tab:Weather Condition}. We varied the parameters Wind Direction, Wind Speed, Turbulence, Gust Speed Increase, and Total Wind Shear to different levels in the high fidelity simulation environment. Figure~\ref{Fig:Simulation II} shows the real-time trajectories for landing in different weather conditions. The initial and final coordinates of the aircraft set to that of the first trial in Table \ref{tab:Different starting positions with engi}. We note that the initial coordinates of the engine-out aircraft are slightly different but relatively close to each other in these simulations because of how we initialize these test runs. As our results suggest, MASC can navigate the aircraft to the configured landing site in each test run under severe weather conditions. Therefore, our approach can robustly plan a landing trajectory and safely navigate the aircraft to a landing site under large wind and turbulence uncertainties. 
    

\begin{table}[ht]
\centering
\begin{tabular}{ | m{2em} | m{2cm}| m{2cm} |m{2cm} |m{2cm} |m{2cm}| } 
  \hline
  Trial & Wind Direction ($deg$) & Wind Speed ($kts$)  & Turbulence ($\%$) & Gust Speed Increase ($kts$) & Total Wind Shear\\ 
  \hline
  1 & 20 & 14 & 10 & 10 & 10  \\ 
  \hline
  2 & 0 & 3 & 8 & 14 & 8      \\ 
  \hline
  3 & 4 & 5 & 10 & 8 & 14     \\
  \hline
  4 & 14 & 7 & 12 & 22 & 8    \\
  \hline
  5 & 27 & 12 & 4 & 9 & 4     \\
  \hline
\end{tabular}
\caption{\small The weather settings in the X-Plane program}
\label{tab:Weather Condition}
\end{table}


\subsection{Feasibility Evaluation for Landing Site Selection}\label{section:simulation_results}

In the MASC framework, the mission adaptation starts with offline trajectory planning for reachability and feasibility evaluation of candidate landing sites. The offline planning architecture, presented in Figure  \ref{Fig:Framework of waypoint follower by software in the loop simulation}, computes feasible emergency landing trajectories for each nearby landing site and selects the best landing coordinates. In our implementation, the control laws and simplified UAV model are implemented in MATLAB/Simulink in acceleration mode. The feasible landing site selection consists of three tasks: offline path planning, landing time estimation, and optimal landing site selection. We build a mathematical model to obtain the discretized reference trajectory points utilized to calculate the landing path and expected landing time for each landing site. To avoid unnecessary calculations, we opt-in for a low density of reference trajectory points (coarser trajectories). The simulations show that the approach is sufficiently fast and computationally inexpensive, and the computation time for each trajectory is $3.083$ s on average (corresponding to a path that takes about $10$ min to complete). However, we note that the computation time depends on the computing hardware used to run these simulations.

\begin{table}[ht]
\centering
\begin{tabular}{ | m{3cm} | m{1.5cm}| m{1.5cm} |m{2cm} |m{2.5cm}|m{1.5cm}|} 
  \hline
  Landing Coordinates & North ($m$) & East ($m$)  & Height ($m$) & Runway Direction ($Deg$) & Landing Time ($s$)\\ 
  \hline
  1 & 21822 & -9751.8 & 235 & 24.17 & 650\\ 
  \hline
  2 & 11822 & -6751.8 & 235 & 130 & 700 \\ 
  \hline
  3 & 46000 & -39751.8 & 235  & 40 & N/A  \\
  \hline
  4 & 36000 & -19751.8 & 235  & 40 & 800  \\
  \hline
\end{tabular}
\caption{\small  Candidate landing coordinates}
\label{tab:Backup Landing Coordinates}
\end{table}
   \begin{figure}[h!]
    \centering
    \includegraphics[width=0.4\textwidth]{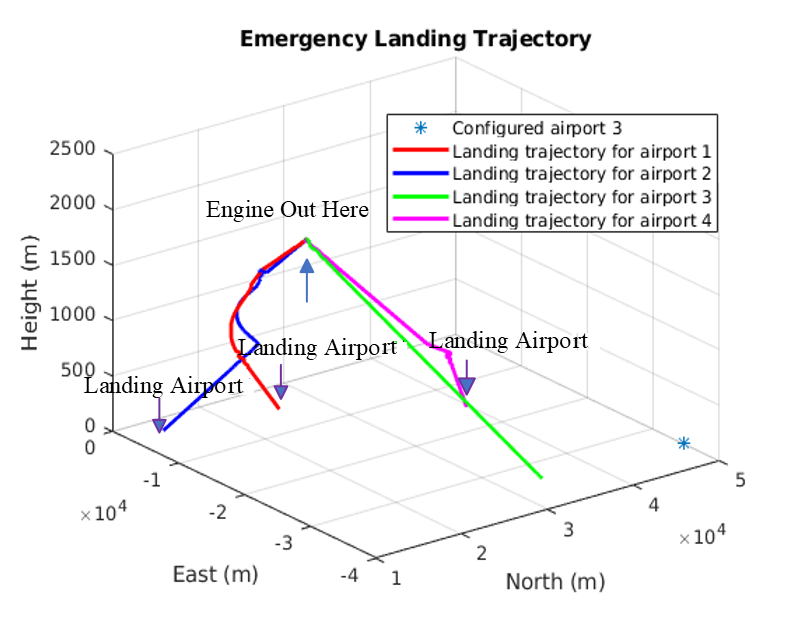}
	\caption{\small Trajectories generated for four different landing sites}
	\label{Fig:Trajectories Planned Offline}
    \end{figure}


In this simulation, our goal is to select the landing site with the shortest landing time among four different potential sites listed in Table~\ref{tab:Backup Landing Coordinates}. Figure~\ref{Fig:Trajectories Planned Offline} shows the trajectories leading to each of the landing sites labeled by 1, 2, 3, and 4, and the corresponding landing times are summarized in Table~\ref{tab:Backup Landing Coordinates}. We note that landing sites 1, 2, and 4 are reachable, but landing site 3 is infeasible, i.e., the engine-out aircraft crashes before reaching the landing site 3. It takes $650$ s to land at landing site 1, which is the shortest time. The emergency landing also takes about 10.08 $min$ in the SITL simulation, which means the offline planning method is accurate enough to predict the actual time needed for the emergency landing process. 



\subsection{Comparison between the Offline and Online Trajectories}

\begin{figure}[h!]
    \centering
    \includegraphics[width=0.4\textwidth]{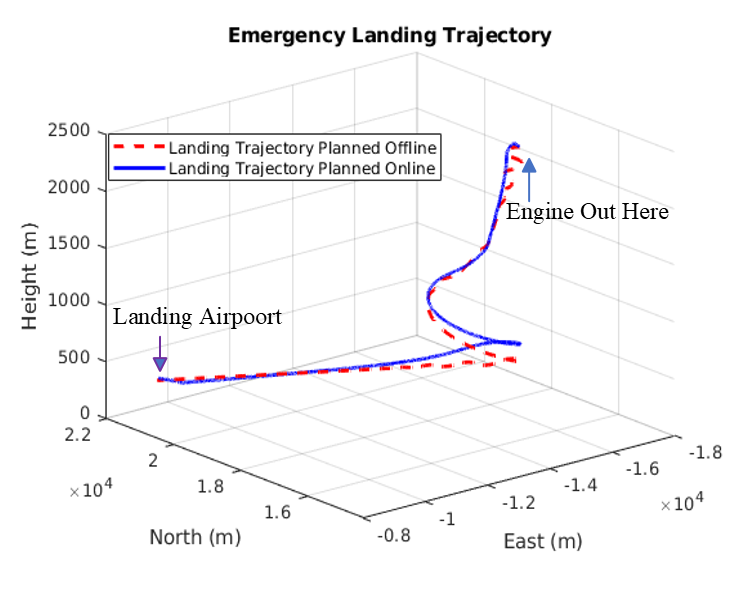}
	\caption{\small Comparison of online and offline trajectories}
	\label{Fig:Online And Offline Trajectory Comparison}
    \end{figure}

We use a simplified UAV model for offline trajectory generation, and therefore, we expect some discrepancies between the predicted trajectory and the actual flight path of the UAV in the high-fidelity simulation. In the final part of our simulation study, we compare the online and offline planned paths to observe possible discrepancies. To do so, we chose the landing site $1$ in SITL simulation and test set for offline path planning. The trajectory planned offline coincides with the online one, as shown in Figure~\ref{Fig:Online And Offline Trajectory Comparison}. Due to fewer turns needed for the aircraft in the SITL simulation to match the heading direction of the reference straight line compared to the offline path planning, the diameter of the path planned offline at the start of Phase I does not entirely match the SITL trajectory. The landing time in the SITL simulation is $6.8$ min, while the estimated landing time is $6$ min for offline path planning in Matlab/Simulink. However, these differences do not considerably impact the main outcomes, which are predicting the feasibility of the selected path and estimation of landing time.

\section{Conclusion}\label{section:conclusion}

This paper proposes a robust and lightweight navigation and control architecture that enables UAVs to perform an automatic emergency landing under severe weather conditions. A Multi-level Adaptation approach in mission planning, path tracking, and stabilizing control was presented within this framework. The proposed framework can also evaluate the feasibility of landing at the nearby landing sites and select the optimal one. Using a high-fidelity simulation environment, the effectiveness of the approach is verified by successfully landing a fixed-wing aircraft under engine failure for a wide range of initial aircraft states and weather uncertainties. In the future, we plan to address other challenges such as vision-based landing and obstacle avoidance when flying over a complex urban area. In this paper, rudder control has not been used. We plan to incorporate rudder control to track a desired angle of sideslip and provide the fixed-wing aircraft with more agility and maneuvering performance.

\section{Acknowledgement}
This research is partially supported by the National Science Foundation (award no. 2137753).

\bibliography{VirtualSully, VirtualSully2}

\end{document}